\RequirePackage{amsmath}
\documentclass[runningheads]{llncs}
\usepackage[T1]{fontenc}
\usepackage{graphicx,verbatim}
\usepackage{placeins}
\usepackage{algorithm}
\usepackage{algpseudocode}
\usepackage{cite}

\newcommand{\figref}[1]{\figurename~\ref{#1}}
\newcommand{\tabref}[1]{\tablename~\ref{#1}}
\newcommand{\secref}[1]{Section~\ref{#1}}

\usepackage{hyperref}

\begin{document}
\title{Beyond one-hot encoding? Journey into compact encoding for large multi-class segmentation}
\titlerunning{Beyond one-hot encoding in large multi-class segmentation?}

\author{
Aaron Kujawa\inst{1} \and
Thomas Booth\inst{1} \and
Tom Vercauteren\inst{1}
}
\authorrunning{A. Kujawa et al.}
\institute{King's College London, London, UK}

\maketitle              \begin{abstract}
This work presents novel methods to reduce computational and memory requirements for medical image segmentation with a large number of classes.
We curiously observe challenges in maintaining state-of-the-art segmentation performance with all of the explored options. 
Standard learning-based
methods typically employ one-hot encoding of class labels.
The computational complexity and memory requirements thus increase linearly with the number of classes.
We propose a family of binary encoding approaches instead of one-hot encoding to reduce the computational complexity and memory requirements to logarithmic in the number of classes. 
In addition to vanilla binary encoding, we investigate the effects of error-correcting output codes (ECOCs),  class weighting, hard/soft decoding, class-to-codeword assignment, and label embedding trees.
We apply the methods to the use case of whole brain parcellation with 108 classes based on 3D MRI images. 
While binary encodings have proven efficient in so-called \emph{extreme classification} problems in computer vision, we faced challenges in reaching state-of-the-art segmentation quality with binary encodings.
Compared to one-hot encoding (Dice Similarity Coefficient (DSC) $= 82.4 (2.8)$), we report reduced segmentation performance with the binary segmentation approaches, achieving DSCs in the range from 39.3 to 73.8.
Informative negative results all too often go unpublished. We hope that this work inspires future research of compact encoding strategies for large multi-class segmentation tasks.


\end{abstract}
\section{Introduction}
Deep learning models for semantic segmentation tasks typically rely on one-hot encoding of class labels. The computational complexity of calculating the class probabilities and the memory required to store them is linear in the number of classes. For the segmentation of high resolution 3D medical images, this linear dependence can lead to excessive resource requirements. While the number of classes for many semantic segmentation tasks in the medical imaging field is often small, an increasing number of applications require a high number of classes. For example, whole brain parcellation (WBP) protocols can include hundreds of distinct classes \cite{cardoso2015geodesic,klein2012101}. 
Another example is TotalSegmentator, a family of foundation models for segmentation of 100+ structures in CT and MRI images \cite{wasserthal2023totalsegmentator,akinci2025totalsegmentator}. 

This has led to the development of mitigation strategies such as patch-based methods which extract smaller sub-volumes at the cost of losing spatial context \cite{moeskops2016automatic,mehta2017brainsegnet,li2017compactness,dolz20183d,wachinger2018deepnat,huo20193d,coupe2019assemblynet,roy20222,mehta2017brainsegnet}. Other approaches reduce these resource requirements by merging related label pairs during model training
and inference \cite{henschel2020fastsurfer,roy2017error,roy2019quicknat,kujawa2024label}.
Model compression, quantization, and related techniques~\cite{li2023model} can also help but the expected gains are not commensurate with the linear class number scaling.

Here, to address the linear increase of resources, we explore a family of different approaches that relies on binary encoding of class labels and thereby frames the original \emph{multi-class} segmentation task as a \emph{multi-task} segmentation problem. 
Each task is a binary segmentation task that corresponds to one bit of the binary representation of the class label. Each voxel will thus typically be associated with several positive bits.
The complexity of this approach is reduced and becomes logarithmic in the number of classes.

To date, binary encoding strategies employed in computer vision have mostly focused on so-called \emph{extreme classification} tasks where large numbers of classes are expected \cite{bengio2019extreme}. Error Correcting Output Codes (ECOCs) have been explored as an efficient way of improving classification performance, improve speed of convergence, and include prior information about the class taxonomy \cite{dietterich1994solving,rodriguez2018beyond,evron2018efficient,evron2023role,radoi2018semantic}.
These works are motivated by the use of lower dimensional output embeddings to encourage the identification and exploitation of intrinsic class relationships. 
Instead, our work focuses on the increased computational and memory requirements associated with a large output space. 

To the best of our knowledge, this work presents the first investigation of binary encoding strategies in the context of semantic 3D medical image segmentation.
Curiously, we find that vanilla binary encoding reduces the segmentation performance compared to one-hot encoding. Our analysis shows that misclassified voxels occur mostly at structure boundaries and predominantly affect small structures. We assess the effect of ECOCs, class weighting, hard and soft decoding of output codes, class-to-codeword assignment, and the use of label embedding trees. While, these strategies did not improve segmentation performance, the negative results presented here serve as a starting point for future research.

\section{Methods and materials}
We use nnU-Net~\cite{isensee2021nnu}, which is recognised as a strong baseline, as a common backbone and implement our proposed output encodings as bespoke final layers with corresponding loss functions.

\subsection{Standard one-hot encoding with multi-class learning}
The standard method of one-hot encoding was used as a reference standard. In the U-Net architecture, the final $1\times1\times1$ convolution maps the feature vectors to $N_\text{C}$ output channels, where $N_\text{C}$ is  the number of classes. A softmax operation yields probability distributions along the class dimension for each voxel.
At training time, the sum of Dice loss and cross-entropy (Dice+CE) loss is applied. The Dice loss is defined as
$
        L_\text{D} = \frac{1}{N_\text{C}} \sum_c{1-\frac{2 \sum_i{p_{c,i}g_{c,i}}}{\sum_i p_{c,i} + \sum_i{g_{c,i}}}}
$
where $p_{c,i}$ is the predicted probability of class $c$ and voxel $i$, and $g_{c, i}$ is the corresponding value of the one-hot encoded ground-truth label.
The cross-entropy loss is defined as
$
    L_\text{X} = \sum_i \frac{1}{\sum_iw_{g_i}} (-w_{g_i}\log(p_{g_i,i}))
$
where the index $g_i$ is the ground truth ordinal class label at voxel $i$ and $w_{g_i}$ is the corresponding class weight.
At inference time, the class with the highest probability is chosen as the predicted output class (argmax operation).

\subsection{Vanilla binary encoding with \emph{multi-label} learning}
This method encodes the class labels in a binary representation. At least $N_\text{B} = \lceil \log_{2}(N_\text{C}) \rceil $ bits are required to represent $N_\text{C}$ classes. For example, $N_\text{B}=7$ binary bits are sufficient to represent up to 128 classes, where each class is represented by a binary number between 0000000 and 1111111.
The CNN's final $1\times1\times1$ convolution maps the feature vector to $N_\text{B}$ output channels, rather than $N_{\text{C}}$ output channels used for standard one-hot encoding. A sigmoid function is applied to each of the output channels to convert output logits to the range (0,1). Each of the output channels is considered as the probabilistic model output of a binary segmentation task, where the output values $p_{c,i}$ represent the probability of the corresponding ground truth bit $g_{c,i}$ to be 1: $p_{c,i} = P(g_{c,i}=1)$. The same loss function (Dice+CE) with $N_\text{C} = 2$ was used to supervise each of the binary output channels.
This method reduces the size of the output tensor size by a factor of $N_\text{C}/N_\text{B}$.

\subsubsection{Hard decoding}
At inference time, hard decoding was employed, i.e., a threshold of 0.5 was applied to the output values $p_{c,i}$ to obtain a binary number. A lookup-table was used to map the binary numbers to the original ordinal class labels. Binary numbers that did not exist in the code book were mapped to the background. 
In the case of vanilla binary encoding, soft decoding as described in \secref{sec:soft_err_corr_dec} is equivalent to hard decoding. Therefore, no separate experiment was conducted.

\subsection{Hamming-based error correcting encoding}
Hamming codes are a family of linear error-correcting codes~\cite{proakis2008digital}. Error-correction is achieved by the introduction of so-called parity bits in addition to the data bits used in vanilla binary encoding. For example, Hamming(7,4) encodes 4 data bits into 7 bits by adding 3 parity bits. 
A generator matrix $\mathbf{G}$ can be used to Hamming(7,4) encode data bits $\mathbf{d} = (d_0, d_1, d_2, d_3)$ as Hamming encoded bits $\mathbf{h} = \mathbf{G}^T \mathbf{d} = (h_1, \dots, h_7)$.
To apply Hamming(7,4) encoding to binary numbers with more than 4 data bits, the binary number can be split into multiple chunks of 4 data bits and each chunk can be encoded and decoded separately. 
The number of required bits is thus at least $N_\text{Hamm} = 7 \lceil \frac{1}{4} \log_{2}(N_\text{C}) \rceil$.
Other ECOCs could be used, for example Hamming(15,7) to avoid chunking. However, here we stick to Hamming(7,4) encoding as the simplest starting point.
The CNN is then trained to predict the encoded bits using a $1\times1\times1$ convolution with one output channel for each encoded bit. At training time, a sigmoid operation is applied to each of the output channels and supervised separately with the corresponding Hamming encoded ground truths using the Dice+CE loss with $N_\text{C}=2$.

\subsubsection{Hard error-correcting decoding}
At inference time, hard error-correcting decoding is achieved by applying a sigmoid operation to the output logits, followed by thresholding at 0.5 to obtain an encoded binary prediction. For each of the 4-bit chunks, error-correcting decoding is performed by calculation of a so-called syndrome vector $\mathbf{z}$, through a modulo 2 binary matrix multiplication with a parity check matrix $\mathbf{P}$, which indicates the erroneous bit. Flipping/negating the erroneous bit of $\mathbf{r}$ results in the corrected Hamming code which can be decoded through another binary matrix multiplication.
Finally, ordinal labels can by restored by means of a look-up table.

\subsubsection{Soft error-correction decoding}
\label{sec:soft_err_corr_dec}
With soft error-correction decoding, the corrected Hamming code is obtained as that Hamming code of the code book with the smallest $L_2$-distance to the model output. With a naive implementation, this approach is computationally expensive and has no memory-saving advantage compared to standard one-hot encoding. It was applied as a means to assess the impact of hard decoding on the segmentation quality knowing that memory efficient soft decoding approaches existing for a range of ECOCs~\cite{proakis2008digital}.

\subsection{Binary-tree encoding, learning, and decoding}
\begin{figure}
\includegraphics[width=\textwidth]{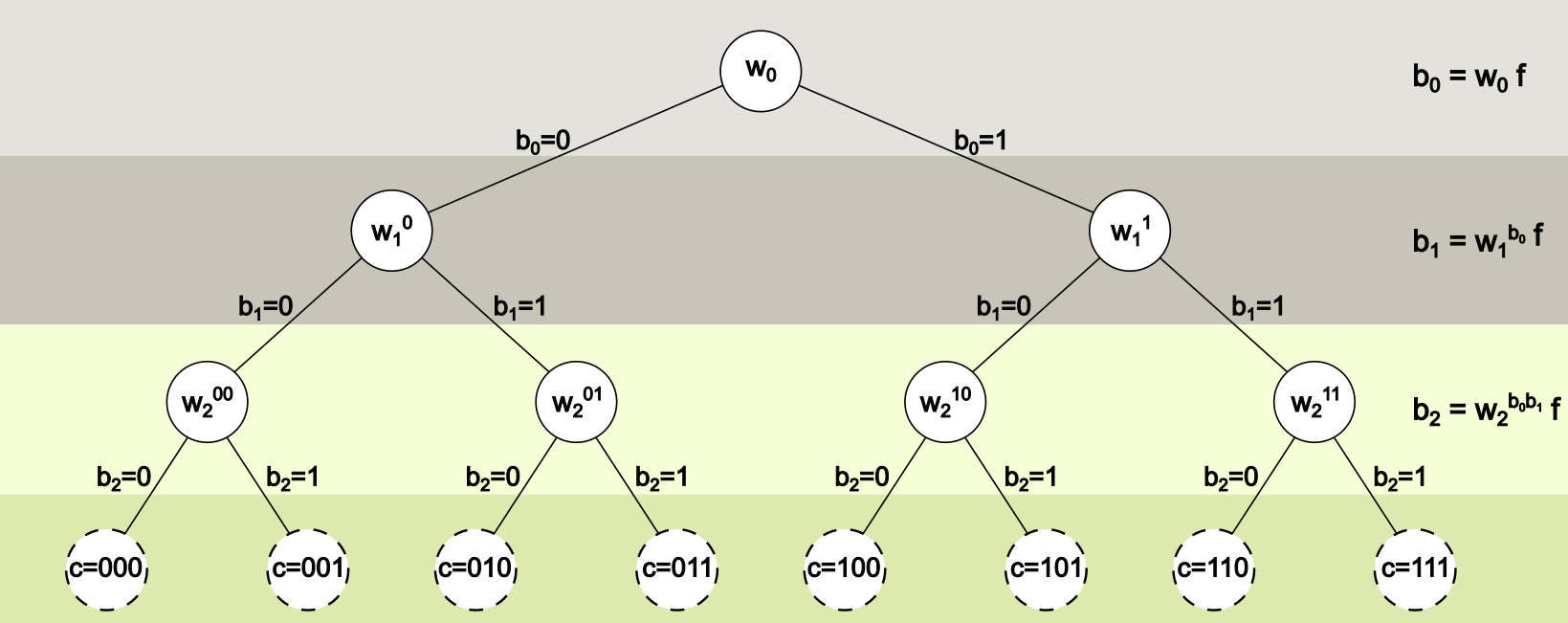}
\caption{Binary tree representing the hierarchy of codeword bits. Weight vectors $\mathbf{w}_i$ for $i>0$ are determined by preceding bits which are determined by the ground truth during training and by the model prediction during inference. The dashed nodes at the bottom of the tree represent the class associated with the path taken from the root node.} \label{fig:binary_tree}
\end{figure}

The method explained in this section and illustrated in \figref{fig:binary_tree} is inspired by \cite{bengio2010label} and aims at increasing the consistency of output channels by conditioning the weights of the $1\times1\times1$ convolution on the bits of the other channels. Specifically, for $N_\text{B} = \lceil \log_{2}(N_\text{C}) \rceil$ output channels with indices $k \in {0,...,N_\text{B}-1}$, the weights $\mathbf{w}_k$ of the $1\times1\times1$ convolution that are associated with output channel $k$ are conditioned on all preceding output channels $\{j | j < k\}$. As shown in \figref{fig:binary_tree}, this approach can be visualized as a binary tree with $N_\text{B}$ levels. While there is only one set of weights for the first output channel, for subsequent channels a different set of weights is applied for each combination of preceding output bits:
\begin{equation}
    \mathbf{w}_k = \begin{cases}
    \mathbf{w}_0 &\text{if } k=0 \\
    \mathbf{w}_k^{b_0,...,b_{k-1}} & \text{otherwise}
    \end{cases}
\label{eq:weight_vecs}
\end{equation}
Hence, for output channel $k$ there are $2^k$ sets of weights, only one of which is selected during each forward pass and updated during the backward pass (training only). 
At training time, the preceding bits $b_0,...,b_{k-1}$ are given by the ground truth, whereas at inference time, $b_0,...,b_{k-1}$ are given by the discretized (sigmoid operation followed by thresholding) network prediction and therefore need to be predicted sequentially.

\subsubsection{Implementation details}
Due to the conditioning on preceding bits, the weights also depend on the spatial location $i$. Typically, implementations of convolutions do not feature a dependence of input weights on spatial location that could be exploited to condition the weights on preceding bits. However, $1\times1\times1$ convolutions are a linear mapping from a feature vector at a single location $i$ to the output vector $b_k(i)$ at that location and can therefore be expressed as a matrix multiplication.
Inclusion of the convolution bias parameter $\mathbf{w}_{k,\text{bias}}$ in the weight parameters $\mathbf{w}_k$ requires appending a 1 at the end of the feature vector $\mathbf{f}$. For matrix multiplication, efficient algorithms exist that allow to incorporate the spatial dependence of the weights: $\mathbf{w}_k \rightarrow \mathbf{w}_k(i)$. Here we apply the \texttt{gather\_mm} function of the Deep Graph Library (DGL) \cite{wang2019deep}.

\subsection{Class to binary code assignment}
The assignment between classes and (encoded) binary numbers is a  bijective mapping called a \textit{code book}. The choice of the code book has been shown to significantly influence performance in classification tasks \cite{evron2023role}. The performance resulting from a code book can depend on 1) the minimum distance $\rho$ between codewords, 2) the degree of correlation between output channels, and 3) the difficulty of the binary sub-problems.
A high minimum distance $\rho$ between codewords leads to better error correction capability.
Low correlation between output channels discourages simultaneous errors in multiple output channels.
A codeword-to-class assignment that leads to binary output maps which are easier to learn can lead to improved performance.
For example, a binary output map with a balanced number of foreground and background voxels may be easier to learn. 
On the other hand, a binary map with multiple foreground structures of complex shapes may present a more difficult sub-problem.

\subsubsection{Random}
As a baseline, a random codeword-to-class assignment was applied. 

\subsubsection{Graph matching based assignment}
The authors of \cite{evron2023role} find that "generalization is superior when encoding similar classes by similar codewords". Following this observation, we employ a graph matching algorithm to optimize a class-to-code assignment with respect to the Hamming distance between similar classes. We consider two classes as similar if we encounter neighboring voxels of those two classes in the training set. This is motivated by the observation that most errors by models trained with random class-to-code assignment are made at the boundary between structures (see \secref{sec:results}). 
Thus, the first graph which represents the similarity between classes is an undirected connected graph with $N_\text{C}$ nodes representing the classes. Two nodes $c_1$ and $c_2$ are connected only if neighboring voxels of both classes are encountered in at least one instance of the training set. The second graph represents the codeword similarity. Each node represents one of $N_{\text{H}} = 2^{N_\text{B}}$ valid codewords, where $N_\text{B}$ is the number of data bits (excluding parity bits). 
Edge weights between nodes are set to 1 if the Hamming distance between codewords is larger than an empirically determined threshold that minimized the Hamming distance between connected classes, otherwise the edge was removed. The pygmtools python library \cite{wang2024pygmtools} was applied to perform the matching between the two graphs.

\subsection{Dataset}
We evaluate the methods on a subset of 585 multi-parametric MRI scans included in the BraTS (Brain Tumor Segmentation) 2021 dataset. Each scan includes 4 modalities: a) native T1-weighted (T1),  b) post-contrast T1-weighted (T1Gd), c) T2-weighted (T2), and d) T2 Fluid Attenuated Inversion Recovery (T2-FLAIR).
The BraTS preprocessing pipeline was replaced with a custom preprocessing pipeline without skull-stripping to prevent the unintentional removal of brain tissue near the skull which is problematic for the whole brain segmentation task. To this end, corresponding unprocessed image volumes were retrieved from The Cancer Imaging Archive (TCIA)~\cite{clark2013cancer}. 
The reference pseudo-ground-truth labels for model training and evaluation were created as a combination of manual segmentations of tumor sub-regions contained in the BraTS 2021 dataset and pseudo-ground truth segmentations generated with the GIF algorithm~\cite{cardoso2015geodesic}.

\section{Results}
\label{sec:results}

Results of all experiments are shown in \tabref{tab:dice}. 
Curiously, compared to traditional one-hot encoding, methods that rely on binary encoding (with and without Hamming error-correction) result in a significant reduction in DSC. Use of the binary tree output head further reduces the DSC. 

\begin{table}[tb]
\footnotesize
    \centering
    \caption{Average Dice Similarity Coefficient (DSC) obtained in all experiments. The reported values are obtained by first calculating the average DSCs over all structures on a case-by-case basis, and then taking the average over all cases. The GPU memory was obtained experimentally for the reported number of output channels by running 1 training epoch with nnU-Net~\cite{isensee2021nnu}.
    \label{tab:dice}}
    \begin{tabular}{|c|c|c|c|c|c|c|c|c|}
    \hline
        \# &
        encoding & 
        loss & 
        head & 
        \begin{tabular}[c]{@{}l@{}}output \\ channels\end{tabular} & 
        decoding & 
        \begin{tabular}[c]{@{}l@{}}class-to-\\ codeword \\ assignment\end{tabular} & 
        avg. Dice &
        \begin{tabular}[c]{@{}l@{}}Mem. \\ {[}GiB{]}\end{tabular} \\
        \hline
         1 & one-hot        & \begin{tabular}[c]{@{}l@{}}multi-channel \\ Dice+CE\end{tabular}
                                    & \begin{tabular}[c]{@{}l@{}} 1x1x1 \\ conv\end{tabular} & \begin{tabular}[c]{@{}l@{}} $N_\text{C} = $ \\ 108 \end{tabular}  
                                    & - &  -                               & 82.4 (2.8) & 26.7 \\ \hline
         2 & vanilla bin.   & \begin{tabular}[c]{@{}l@{}}binary \\ Dice+CE\end{tabular}
                                    & \begin{tabular}[c]{@{}l@{}} 1x1x1 \\ conv\end{tabular} & \begin{tabular}[c]{@{}l@{}} $N_\text{B} = $ \\ 7 \end{tabular}
                                    & hard & random                           & 72.7 (3.3) & 12.08 \\ \hline
         3 & Hamming  & \begin{tabular}[c]{@{}l@{}}binary \\ Dice+CE\end{tabular}
                                    & \begin{tabular}[c]{@{}l@{}} 1x1x1 \\ conv\end{tabular} & \begin{tabular}[c]{@{}l@{}} $N_\text{H} = $ \\ 14 \end{tabular}
                                    & hard &  random                       & 73.7 (3.2) & 12.21 \\ \hline
         4 & Hamming  & \begin{tabular}[c]{@{}l@{}}weighted \\ binary CE\end{tabular}
                                    & \begin{tabular}[c]{@{}l@{}} 1x1x1 \\ conv\end{tabular} & \begin{tabular}[c]{@{}l@{}} $N_\text{H} = $ \\ 14 \end{tabular}
                                    & hard &  random                       & 71.9 (3.3) & 12.21 \\ \hline
         5 & Hamming  & \begin{tabular}[c]{@{}l@{}}weighted \\ binary CE\end{tabular}
                                    & \begin{tabular}[c]{@{}l@{}} 1x1x1 \\ conv\end{tabular} & \begin{tabular}[c]{@{}l@{}} $N_\text{H} = $ \\ 14 \end{tabular}
                                    & soft &  random                       & 73.8 (3.3) & - \\ \hline
         6 & Hamming  & \begin{tabular}[c]{@{}l@{}}weighted \\ binary CE\end{tabular}
                                    & \begin{tabular}[c]{@{}l@{}} 1x1x1 \\ conv\end{tabular} & \begin{tabular}[c]{@{}l@{}} $N_\text{H} = $ \\ 14 \end{tabular}
                                    & hard &   
                                    \begin{tabular}[c]{@{}l@{}}graph- \\ matching\end{tabular}                                                      
                                                                           & 72.4 (2.9) & 12.21 \\ \hline
         7 & vanilla bin.  & \begin{tabular}[c]{@{}l@{}}binary \\ Dice+CE\end{tabular}
                                    & tree                                                   & \begin{tabular}[c]{@{}l@{}} $N_\text{B} = $ \\ 7 \end{tabular}
                                    & hard &  random                       & 39.3 (1.5) & - \\ \hline
    \end{tabular}
\end{table}

We note that one-hot encoding outperforms the other encoding strategies significantly for small structures such as \textit{inferior lateral ventricles} or \textit{vessels} while the differences in DSC for larger structures such as \textit{cerebellum} or \textit{cerebral white matter} are small, except for the binary-tree encoding strategy which significantly under-performs for the majority of classes.
\begin{figure}[tbh!]
\includegraphics[width=\textwidth]{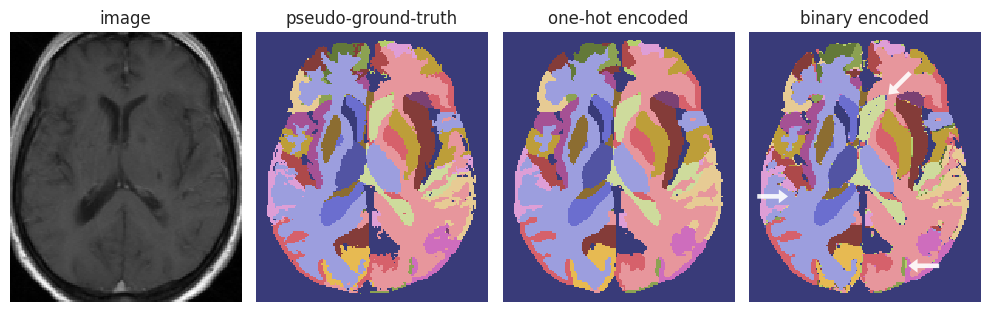}
\caption{Comparison of predicted segmentations with one-hot encoding and vanilla binary encoding. Both encoding strategies yield sensible predictions. However, the white arrows point out examples of misclassified pixels at structure boundaries as a result of the binary encoding strategy.} \label{fig:slices}
\end{figure}
\begin{figure}[tbh!]
\includegraphics[width=\textwidth]{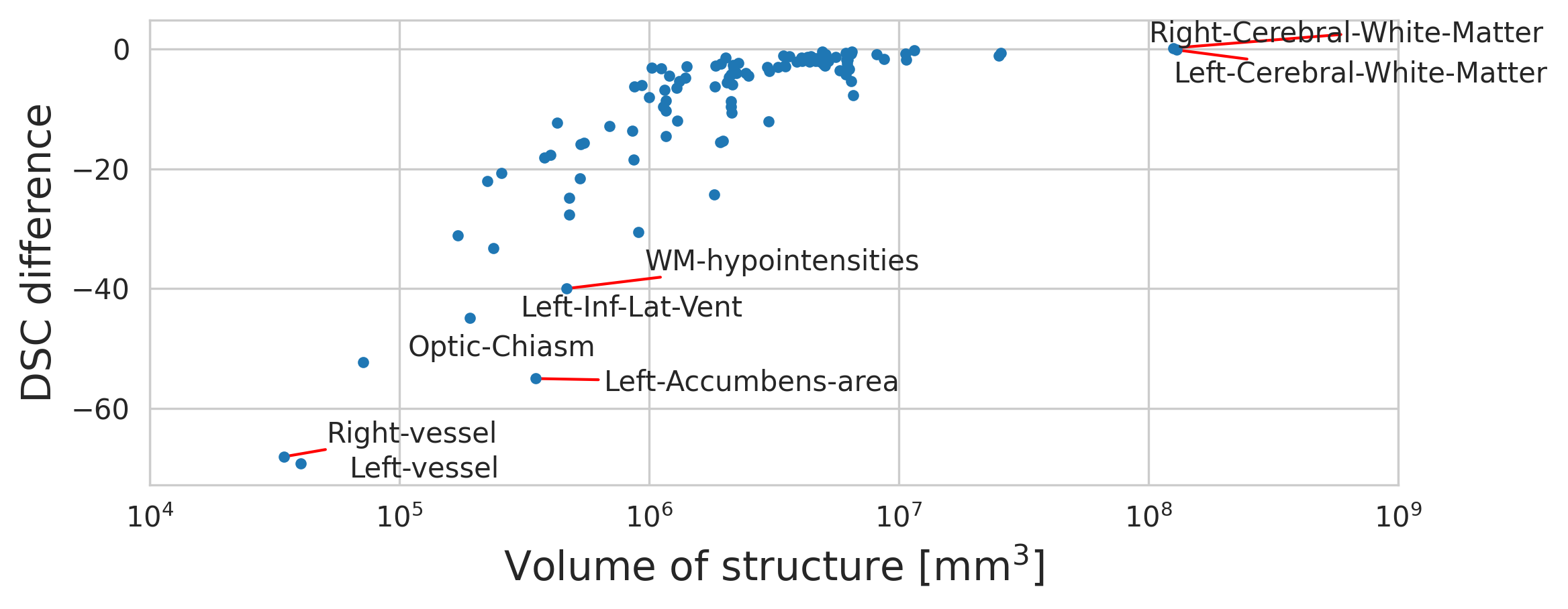}
\caption{Comparison between one-hot encoding and vanilla binary encoding. The difference in average DSC between both encoding strategies shows that small structures experience the largest drop in DSC.} \label{fig:dscdrop}
\end{figure}
\figref{fig:slices} shows a comparison of label map predictions with one-hot encoding and vanilla binary encoding and highlights the main binary encoding artifact that affects voxels at structure boundaries. 
\figref{fig:dscdrop} shows that classes of small volume have significantly lower DSC when binary encoding is used, while classes of larger volume are less affected. Use of a weighted binary CE loss function (experiments 4-6 of \tabref{tab:dice}) did not lead to significant improvements.

\section{Discussion}
This work investigates the impact of different binary encodings on segmentation performance and associated requirements. 
Despite positive results published in the computer vision literature,
we find that, when applied to medical image segmentation, these strategies lead to decreased performance compared to one-hot encoding. Attempts to improve the segmentation performance with ECOCs, spatial weighting, soft decoding, similarity preserving class-to-codeword assignments, or predictions conditioned on preceding output bits did not yield significant improvements over the vanilla binary encoding strategy. 

Our analysis shows that binary encoding results in a higher number of mis-classifications in the vicinity of structure boundaries. In terms of DSC, small structures are affected the most. These mis-classifications are a result of the independence of predictions made by each of the bit-wise output heads. 
The binary tree approach presented in this work was applied to reduce the mis-classifications by conditioning the weights of the final CNN layer on the predictions of previous output heads. However, a further DSC reduction was observed.
Although the binary tree approach did not yield satisfactory segmentation performance, we speculate that a method that enforces consistency between predicted output heads is required to close the segmentation quality gap with one-hot encoding. 

We hope that our informative negative results can serve as a stepping stone for the community to develop novel strategies for large multi-class segmentation.

\FloatBarrier

\begin{credits}
\subsubsection{\ackname}
This work was supported by the MRC [MR/X502923/1] and core funding from the Wellcome/EPSRC [WT203148/Z/16/Z; NS/A000049/1].
For the purpose of open access, the authors have applied a CC-BY public copyright license to any Author Accepted Manuscript version arising from this submission.

\subsubsection{\discintname}
TV is co-founder and shareholder of Hypervision Surgical.
The authors have no other relevant interests to declare.
\end{credits}

\bibliographystyle{splncs04}
\bibliography{EMA4MICCAI2025-Kujawa}
\end{document}